\documentclass[]{easychair} 
\usepackage{graphicx} 

\usepackage{todonotes}
\usepackage{reledmac}
\usepackage{paralist}
\usepackage[utf8]{inputenc} 
\usepackage[T1]{fontenc}    
\usepackage{booktabs}       
\usepackage{amsfonts}       
\usepackage{nicefrac}       
\usepackage{microtype}      
\usepackage{booktabs} 
\usepackage{subcaption}
\usepackage{graphicx} 
\usepackage{pgfplots} 
\pgfplotscreateplotcyclelist{rw}
{solid, mark repeat = 1, mark phase = 1, mark = *, black\\
	solid, mark repeat = 1, mark phase = 1, mark = square*, black\\}

\usepackage{tikz} 
\usetikzlibrary{shapes,arrows}
\usepackage{pgfplots}
\usepackage{bussproofs}
\usepackage{xspace}
\usepackage{amsmath}
\usepackage{amssymb}
\usepackage{mathtools}

\def\sml{\textsf{SML}\xspace}

\def\sml{\textsf{Standard ML}\xspace}
\usepackage{color}

\usetikzlibrary{arrows,shapes,calc}
\usetikzlibrary{trees,positioning,fit}
\tikzstyle{arrow}=[draw,-to,thick]
\tikzstyle{embedding} = [draw, minimum width=8mm, minimum height=6mm]
\tikzstyle{nnop} = [draw, minimum width=8mm, minimum height=8mm, rounded 
corners]
\tikzstyle{block} =
[rectangle, draw,
minimum width=9mm, text centered, rounded corners, minimum height=2.0em]
\tikzstyle{smallblock} =
[rectangle, draw,
minimum width=9mm, text centered, minimum height=2.0em]

\tikzstyle{line}=[draw]
\tikzstyle{cloud} =
[draw, text centered, ellipse, minimum height=2.0em, minimum width=9em]

\newcommand{\mloop}{\mathit{loop}}
\newcommand{\mloopt}{\mathit{loop2}}
\newcommand{\mcompr}{\mathit{compr}}
\newcommand{\mcond}{\mathit{cond}}
\newcommand{\mdiv}{\mathit{div}}
\newcommand{\mmod}{\mathit{mod}}
\newcommand{\mathleft}{\@fleqntrue\@mathmargin0pt}

\title{Learning Program Synthesis for Integer Sequences\\ from Scratch}
\author{Thibault Gauthier, Josef Urban}
\institute{Czech Technical University in Prague, Prague, Czech Republic\\
\email{email@thibaultgauthier.fr, josef.urban@gmail.com}}
\titlerunning{Learning Program Synthesis for Integer Sequences}
\authorrunning{T. Gauthier and J. Urban}

\begin{document}

\maketitle

\begin{abstract}
We present a self-learning approach for synthesizing programs
from integer sequences. Our method relies on a tree search guided by a learned 
policy. 
Our system is tested on the On-Line Encyclopedia of Integer Sequences. There, 
it discovers, on its own, solutions for 27987 
sequences starting from basic 
operators and without human-written training examples.
\end{abstract}

\section{Introduction}
The search for abstract patterns is one of the principal occupations of
mathematicians. The discovery of similar patterns across different mathematical 
fields often leads to surprising connections. Probably the most famous example of such 
an unexpected connection in mathematics is the Taniyama–Shimura 
conjecture proved in 2001~\cite{breuil2001modularity}. It relates elliptic 
curves over the 
field of rational numbers with a special kind of complex 
analytical functions known as modular forms. This conjecture became especially 
famous because a restricted version of it implies Fermat's last 
theorem~\cite{wiles1995modular}.
The connections found by the system described in this paper are more modest.
For instance, it has created formulas for 
testing prime numbers based both on Fermat's little theorem\footnoteA{\url{https://en.wikipedia.org/wiki/Fermat_primality_test}} (stated in 1640) and Wilson's 
theorem\footnoteA{\url{https://en.wikipedia.org/wiki/Wilson's_theorem}} (stated in 1770).
When a symbolic representation (e.g. formula) describing a pattern is 
conjectured and preferably proven, a mathematician can start reasoning and 
deriving additional facts about the
theory in which the pattern occurs. 

Integer sequences are a very common kind of mathematical 
patterns. A compilation of such sequences
is available in the On-Line Encyclopedia of Integer Sequences (OEIS)~\cite{oeis}. 
Our objective in this project is 
to create a system that can 
discover, on its own, programs for the OEIS sequences.
Such programs will be chosen to be small since usually the best explanations for a 
particular phenomenon are the simplest ones. 
This principle is known as \emph{Occam's razor} or the \emph{law of parsimony}. 
It has been one of the most important heuristics guiding scientific research in general.
In a machine learning setting, this can be seen as a form of regularization.
A mathematical proof of this principle relying on Bayesian 
reasoning and assumptions about the computability of our universe is 
provided in Solomonoff's theory of inductive 
inference~\cite{SOLOMONOFF19641}.

The programs produced by our system are readable, making it possible 
for mathematicians to gain insight into the nature of the patterns 
by analyzing the programs.\footnoteA{See our web interface~\cite{oeis-synthesis-web}.}
We believe that in the future, such systems will assist
mathematicians during their conjecture-making process when investigating
open problems.

\subsection{Overview}
Our approach is to synthesize programs for the OEIS sequences (Section~\ref{sec:oeis}) in a simple domain-specific
language (Section~\ref{sec:prog}) in many iterations of the following 
self-learning loop (Figure~\ref{fig:code1}).
Each iteration 
(generation) of the 
loop consists of
three phases: a generating phase (Section~\ref{sec:generate}), 
        a testing phase
(Section~\ref{sec:test})
and a training phase 
(Section~\ref{sec:train}).
Initially, multiple searches 
are randomly building programs that generate integer sequences and checking if they are in the OEIS.
Then, 
for each generated OEIS sequence we 
select the smallest program that generates it.
From those solutions, a tree neural network is trained to predict what
the right building action 
is, given a target sequence and a partially built 
program. 
The next searches are then guided by the statistical correlations learned by 
the network, typically producing further solutions.
We describe the experiments in Section~\ref{sec:experiments} and analyze some of the solutions in Section~\ref{sec:analysis}.
Section~\ref{sec:related} discusses related work.
\begin{figure}[]
  \begin{small}
    \begin{center}
        \begin{minipage}[b]{0.8\linewidth}
\begin{verbatim}



tnn = random_initialization ();
while(true)	
{
  programs = generate (tnn, sequences);
  solutions = test (programs, oeis);
  tnn = train (solutions)
}
\end{verbatim}
        \end{minipage}
        \caption{\label{fig:code1}Pseudo-code of the self-learning procedure.}
    \end{center}
  \end{small}
\end{figure}

\section{The OEIS}
\label{sec:oeis}

As of March 2022, the number of OEIS entries reached 351663.
All our experiments and analysis are performed on this version.
The sequences come from various mathematical domains such as combinatorics, 
number theory, geometry, and group 
theory. Each sequence is accompanied by its description in English. 
Sequences in the OEIS are expected to be infinite, but for obvious reasons, only 
a finite number of terms are provided. Sequences are only accepted by the 
OEIS curators 
if enough terms are provided according to their judgment.
The distribution of sequences with respect to their length in the OEIS is given 
in Figure~\ref{fig:len}\ .

\pgfplotscreateplotcyclelist{rw}
{solid, mark repeat = 10, mark phase = 200, mark = *, black\\
	solid, mark repeat = 1, mark phase = 1, mark = square*, black\\}
\pgfplotsset{scaled y ticks=false}
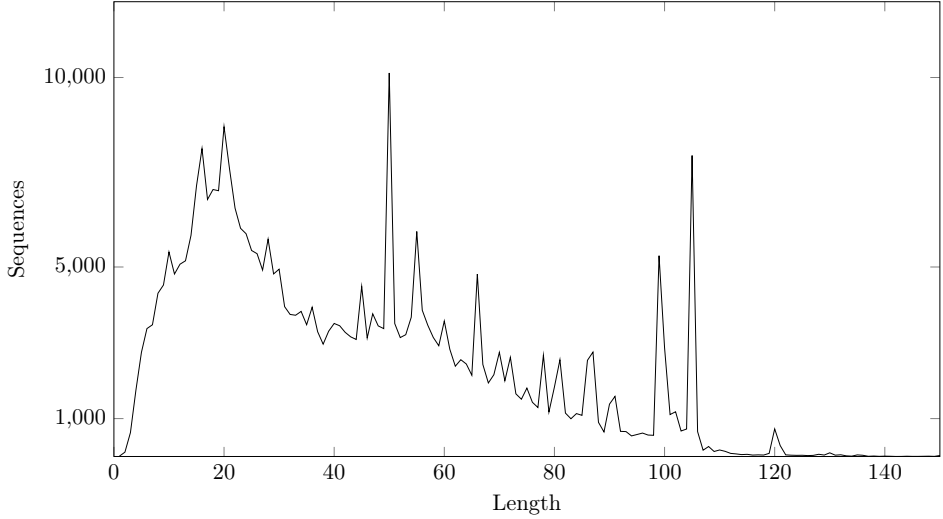
\begin{figure}[]
	\begin{tikzpicture}[scale=0.85]
	\begin{axis}[
	width=1*\linewidth,
	height=0.6*\linewidth,xmin=0, xmax=150,
	ymin=0, ymax=12000,
	ytick={1000,5000,10000},
	cycle list name=rw,
        ylabel near ticks,
	xlabel near ticks,
	xlabel = {Length},
	ylabel = {Sequences} 
	]
	\addplot table[x=len, y=occ] {length};
	\end{axis}
	\end{tikzpicture}
	\caption{\label{fig:len}Number $y$ of OEIS sequences with length $x$.}
\end{figure}

\section{Programming language}\label{sec:prog}
To limit the influence of our human knowledge about the OEIS, we include in our domain-specific
language only several basic arithmetical constants and operators such as $0, +, \mdiv$, and fundamental programming 
constructs such as variables and loops. Informally, the final results of our 
construction are programs that implement functions mapping integers to 
integers ($\mathbb{Z}\mapsto \mathbb{Z}$) built from such operators and 
constructs.
Higher-order arguments of looping operators are 
binary functions $(\lambda(x,y).p: \mathbb{Z}^2 \mapsto 
\mathbb{Z}$). There, the sub-expression constituting the subprogram $p$
may depend on both $x$ and $y$.
Formally, the set $P$ of programs and subprograms in our language, together with 
the auxiliary set $F$ of binary functions (higher-order arguments), 
are inductively defined to be the smallest sets such that
$0,1,2,x,y\in P$,
and if $a,b,c\in P$ and  $f,g \in F$
then:
\begin{align*}
  &a + b, a - b, a \times b, a\ \mdiv\ b, a\ \mmod\ b, 
\mcond(a,b,c) \in P, \\
 &\lambda(x,y).a \in F, \\ 
 &\mloop(f,a,b), \mloopt (f,g,a,b,c), \mcompr(f,a) \in P
\end{align*}
The interpretations of $0,1,2$ are the corresponding constant functions on $ 
\mathbb{Z}^2 \mapsto \mathbb{Z}$, while $x$ and $y$ 
are the two projections in $\mathbb{Z}^2 \mapsto \mathbb{Z}$.
All the other operators and constructs follow the semantics of \sml \cite{harper1986standard} except for 
$\mcond,\mloop,\mloopt,\mcompr$. 
Given subprograms $a,b,c \in \mathbb{Z}^2\mapsto\mathbb{Z}$ and functions $f,g 
\in \mathbb{Z}^2 \mapsto \mathbb{Z}$, we give the following definitions for the 
semantics of 
those operators:\footnoteA{To shorten the notation, we write here
  just $a$, $a-1$, and $\mcond (a,b,c)$ when the argument(s) are some implied 
  input(s) $x_0$ ($y_0$), i.e., these stand for 
  $a(x_0,y_0)$, 
  $a(x_0,y_0)-1(x_0,y_0)$, and  $\mcond (a(x_0,y_0),b(x_0,y_0),c(x_0,y_0))$.}

\begin{align*}
&\mcond (a,b,c) :=\ \mbox{if } a \leq 0 \mbox{ then } b \mbox{ else } c\\
&\mloop (f,a,b) :=\ b \mbox{ if } a \leq 0\\
&\hspace{5mm} f (\mloop(f,a-1,b),a) \mbox{ otherwise}\\
&\mloopt (f,g,a,b,c) :=\ b \mbox{ if } a \leq 0\\
&\hspace{5mm}\mloopt (f,g,a-1,f(b,c),g(b,c)) \mbox{ otherwise}\\
&\mcompr (f,a) :=\ \mbox{failure\ if } a < 0\\
&\mathit{min} \lbrace m\ |\ m \geq 0 \wedge f(m,0) 
\leq 0 \rbrace 
\mbox{ if } a = 0\\
&\mathit{min} \lbrace m\ |\ m > \mcompr (f,a-1) 
\wedge f(m,0) 
\leq 0 \rbrace \mbox{ otherwise}\\
\end{align*}

The program $\mcompr(f,x)$ constructs all the elements $m$ satisfying the 
condition $f(m,0) \leq 0$ as $x$ increases. That is why we call it $\mcompr$
which is a shorthand for set \emph{comprehension}.
In theory, the presence of the minimization operator $\mcompr$ guarantees that
this language is Turing-complete. In practice, the expressiveness of the 
language largely depends on the allocated time and memory. 
The two other looping operators $\mloop$ and $\mloopt$ can also be defined 
with recurrent relations:
\begin{align*}
&\mloop (f,a,b) := u_a\ \mbox{where}\ u_0 = b, u_n = f(u_{n-1},n)\\
&\mloopt (f,g,a,b,c) := u_a\ \mbox{where}\ (u_0,v_0) = (b,c)\ \mbox {and}\\
&\hspace{5mm} (u_n,v_n) = (f(u_{n-1},v_{n-1}), g(u_{n-1},v_{n-1}))\\
\end{align*}

The size of a program is measured by counting the number of operators composing 
it. This number can be obtained by counting the number of tokens in the 
expression ignoring binders ($\lambda(x,y).$), commas and parentheses. 
For example, the program $\mloop (\lambda(x,y).\ x \times y, x, 1)$ has size 6.

\section{Generating programs} \label{sec:generate}
To generate programs, we rely on multiple searches.
Each of them targets a randomly chosen OEIS sequence $s$ and constructs a set 
of programs intended to generate $s$. 
First, we describe our bottom-up process for constructing a 
single program. Then, we explain how this method can be extended to 
synthesize multiple programs at once by sharing construction nodes in a 
policy-guided tree search.

\paragraph{Program construction}
Programs are built following their reverse polish notation. 
Starting from an empty stack, a learned policy is used to repeatedly choose the 
next operator to push on top of a stack. This policy
is computed by a tree neural network (Section~\ref{sec:train}) based on the 
target sequence $s$ and the current stack $k$. 
For instance, the program $\mloop (\lambda(x,y).\ x \times y, x, 1)$ can be 
built from the empty stack by the following sequence of actions:
\[[\ ]  \rightarrow_x [x] \rightarrow_y [x,y] \rightarrow_\times  [x \times y] 
\rightarrow_x [x \times y, x] \rightarrow_1\] \[[x \times y, x, 1]
\rightarrow_\mloop [\mloop (\lambda(x,y).\ x \times y, x, 1)]\]

The stack here starts as an empty list, growing up to length 3 after the fifth action. Then
the final $\mloop$ action reduces it to a stack of length 1
containing the final program.
Note that the higher-order actions such as $\mloop$ include
the creation of $\lambda$-terms (here $\lambda(x,y).\ x \times y$) as the appropriate 
functional arguments. 

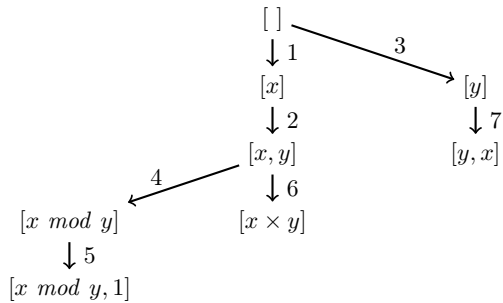
\begin{figure}[t]
	\centering
	\begin{tikzpicture}[scale=0.9,every node/.style={scale=0.9},node 
	distance=1cm]
	\node [] (1){$[\ ]$};
    \node [below of=1] (2) {$[x]$};
	\node [right of=2, node distance=3cm] (2r) {$[y]$};
    \node [below of=2] (3) {$[x,y]$};
	\node [right of=3, node distance=3cm] (3r) {$[y,x]$};
	\node [below of=3] (4) {$[x \times y]$};
	\node [left of=4, node distance=3cm] (4l) {$[x\ \mmod\ y]$};
	\node [below of=4l] (5) {$[x\ \mmod\ y,1]$}; 
	\draw[-to,thick] (1) to node[xshift=3mm] {1} (2); 
	\draw[-to,thick] (1) to node[xshift=4mm,yshift=1mm] {3} (2r); 
	\draw[-to,thick] (2) to node[xshift=3mm] {2} (3);
	\draw[-to,thick] (3) to node[xshift=3mm] {6} (4);
	\draw[-to,thick] (3) to node[xshift=-4mm,yshift=1mm] {4} (4l);
	\draw[-to,thick] (4l) to node[xshift=3mm] {5} (5);
    \draw[-to,thick] (2r) to node[xshift=3mm] {7} (3r);
	\end{tikzpicture}
	\caption{\label{fig:search}
	 7 iterations of the search loop gradually extending the search tree.
     The iteration number leading to the creation of a given node/stack is 
     indicated on the arrow/action leading to it. The set of the synthesized 
     programs after the 7th iteration is $\lbrace 1,x,y,x \times y,x\ \mmod\ y
     \rbrace$.
     }
\end{figure}

\paragraph{Tree search}
The search process is initialized with an empty stack at the root of our 
\emph{search tree} (Figure~\ref{fig:search}).
Starting from this root, actions are selected 
according to the action probabilities returned by the \emph{policy predictor} (Section~\ref{sec:train}) for the 
current node until an action whose output does not yet appear in the tree is 
selected.
Then, a node is created containing the result of that action.
After the creation of a new node, the selection process is restarted from the 
root. The search is run until a given timeout and we collect all the programs 
generated during the search for the next phase (see Section~\ref{sec:test}).
To speed up the search, we cache embeddings (see Section~\ref{sec:train}) for 
all sub-trees in the target 
sequence and in the program stack.
An illustration of the gradual extension of the search tree is given in 
Figure~\ref{fig:search}.

\section{Testing programs}\label{sec:test}
To see if the synthesized programs produce OEIS sequences, we evaluate (test) them on an initial segment of natural numbers.
To do that, we first remove all (potentially) binary programs, i.e., those where $y$ occurs free (not under $\lambda$),
from the set of programs collected during the search(es).
Then, for each remaining program $p$, we generate its sequence 
$p(0),p(1),p(2),\ldots$.
Formally, we say that a program $p$ \emph{covers} (or \emph{is a solution for}) a finite 
sequence $(s_x)_{0\leq x\leq n}$ if 
and only if:
\[\forall x\in \mathbb{Z}.\ 0 \leq x \leq n \Rightarrow p(x) = s_x\]
As an example, the program $\mloop (\lambda(x,y).\ x \times y, x, 1)$ 
(i.e. $u_x$ where $u_0 = 1$ and $u_n = u_{n-1} \times n$)
covers any finite sequence of factorial numbers (OEIS sequence
A142\footnoteA{The sequence is at \url{https://oeis.org/A000142} . Its program
  can be found with our web interface~\cite{oeis-synthesis-web}
and input $1,1,2,6,24$ at
\url{http://grid01.ciirc.cvut.cz/~thibault/cgi-bin/qsynt_v3.cgi?Formula=1,1,2,6,24}.}).
 
Finally, for each OEIS sequence $s$, we select the shortest program 
$p$ among its 
solutions. If there are multiple programs with the smallest size, we break the 
ties using a fixed total order.
After that, training examples are extracted from such selected sequence-solution pairs,
see Section~\ref{sec:train}\ .

Note that during the search targeting a particular sequence $s$, programs for 
many other OEIS sequences may be generated. These programs can be considered as positive examples together with
their respective OEIS sequences. This process is a form of
\emph{hindsight experience replay}~\cite{andrychowicz2017hindsight} where the failed 
attempts that lead to other OEIS targets are used for training.

\section{Training from solutions}\label{sec:train}

A tree neural network (TNN)~\cite{GollerK96} is used as our policy predictor since its dynamic 
structure naturally matches the tree structure of our programs.
This machine learning model is both sufficiently fast for our purpose and has been shown to perform 
well on several arithmetical tasks~\cite{DBLP:conf/mkm/Gauthier20}, outperforming
several other models including the 
NMT~\cite{luong2015effective}
recurrent neural toolkit.


\paragraph{Definition}
A TNN projects labeled trees into an embedding space $\mathbb{R}^{d}$.
Given a tree $t=f(t_1,\ldots,t_n)$, an 
embedding function  $e: \mathit{Tree} \mapsto 
\mathbb{R}^d$ can be recursively defined by:
\[e(f(t_1,\ldots,t_n))=_\mathit{def} N_f(e(t_1),\ldots,e(t_n))\]
where $N_f$ is a function from $\mathbb{R}^{n \times d}$ to $\mathbb{R}^d$
computed by the neural network building block associated with the 
operator $f$.

\paragraph{Training examples}
The basis for creating the set of the training examples are the
covered OEIS sequences and their shortest discovered
programs. In general, each sequence $s$ and its program $p$ generate
multiple training examples corresponding to the states and actions
involved in building $p$.

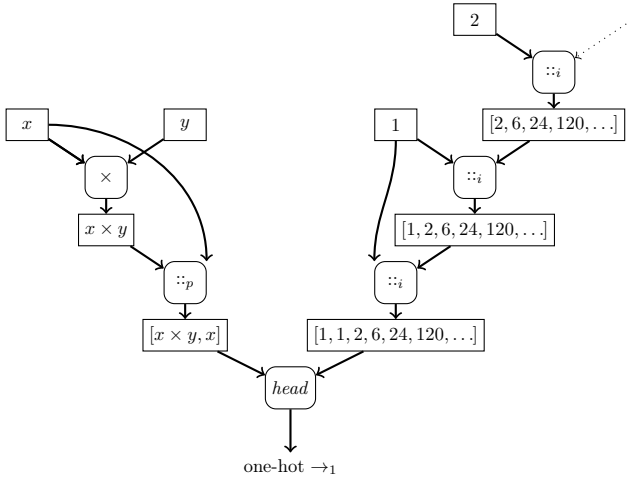
\begin{figure}[t]
	
	\begin{tikzpicture}[scale=0.7,every node/.style={scale=0.7},node 
	distance=3cm]
	
	\node [embedding] (two) {$x$};
	\node [right of=two,node distance=1.5cm] (n+1I) {};    
	\node [embedding,right of=two] (I) {$y$};     
	\node [nnop,below of=n+1I,node distance=1cm] (sum) {$\times$};    
	\node [embedding,below of=sum,node distance=1cm] (sumr) {$x \times 
		y$};   
	\node [right of=sumr, node distance=1.5cm] (2m) {};
	\node [nnop,below of=2m,node distance=1cm] (times2) 
	{$::_p$};      
	\node [embedding,below of=times2,node distance=1cm] (times2r) {$[x \times 
		y,x]$};
	\node [right of=times2r,node distance=2cm] (times2rm) {};        
	\node [embedding, node distance=4cm, right of=times2r] (times2rr) 
	{$[1,1,2,6,24,120,\ldots]$};
	\node [nnop,below of=times2rm,node distance=1cm] (eq) 
	{$\mathit{head}$};   
	\node [below of=eq,node distance=1.5cm] (eqr) {one-hot $\rightarrow_1$}; 
	
	\node [nnop, node distance=1cm, above of=times2rr] (catq) {$::_i$};  
	\node [node distance=1cm, above of=catq] (catqm) {};     
	\node [embedding,node distance=1.5cm, right of=catqm] (catqr) 
	{$[1,2,6,24,120,\ldots]$}; 
	\node [nnop, node distance=1cm, above of=catqr] (catq2) {$::_i$};  
	\node [node distance=1cm, above of=catq2] (catq2m) {};     
    \node [embedding,node distance=1.5cm, left of=catq2m] (catq2l) {$1$};  
	\node [embedding,node distance=1.5cm, right of=catq2m] (catq2r) 
	{$[2,6,24,120,\ldots]$}; 
	
    \node [nnop, node distance=1cm, above of=catq2r] (catq3) {$::_i$};  
	\node [node distance=1cm, above of=catq3] (catq3m) {};     
	\node [embedding,node distance=1.5cm, left of=catq3m] (catq3l) {$2$};  
	\node [node distance=1.5cm, right of=catq3m] (catq3r) 
	{}; 
	\draw[-to,thick] (two) to (sum);
    \draw[-to,thick] (two) to [out=0,in=90] (times2.north east);
    \draw[-to,thick] (two) to (sum);
	\draw[-to,thick] (I) to (sum);
	\draw[-to,thick] (sum) to (sumr);
	\draw[-to,thick] (sumr) to (times2);  
	\draw[-to,thick] (times2) to (times2r);  
	\draw[-to,thick] (times2r) to (eq);  
	\draw[-to,thick] (times2rr) to (eq);  
	\draw[-to,thick] (eq) to (eqr);
	\draw[-to,thick] (catq) to (times2rr);
	\draw[-to,thick] (catqr) to (catq);
	\draw[-to,thick] (catq2) to (catqr);
	\draw[-to,thick] (catq2r) to (catq2);
	\draw[-to,thick] (catq2l) to (catq2);
	\draw[-to,thick] (catq2l) to [out=270,in=90] (catq.north west);
	\draw[-to,thick] (catq3) to (catq2r);
	\draw[-to,dotted] (catq3r) to (catq3);
	\draw[-to,thick] (catq3l) to (catq3);
	\end{tikzpicture}
	\caption{\label{fig:tnn}
		Rectangles represent embeddings and 
		rounded squares represent neural network blocks.
	    The operator $::_p$ (respectively $::_i$) is a list constructor for 
	    program stacks (respectively integer sequences).}

\end{figure}

In more detail, each such training example consists of a pair $\mathit{(head(k,s),\ 
\rightarrow_a)}$.
The term $\mathit{head}(k,s)$ joins a stack $k$ and 
a 
sequence $s$ into a single tree with the help of an 
additional 
operator $\mathit{head}$.
And $\rightarrow_a$ is the action required on $k$ to progress towards the 
construction of $p$.
From the solution $\mloop (\lambda(x,y).\ x \times y, x, 1)$ (Figure~\ref{fig:tnn})
we can extract one training example per action leading to its creation. One 
such example is $((\mathit{head}([x \times y,x],[1,1,2,6\ldots]), \ 
\rightarrow_1)$. Our TNN is actually predicting a 
policy distribution over all the actions. So, the action $\rightarrow_1$ is 
projected to $\mathbb{R}^d$ using a one-hot embedding. This essentially tells 
the network that every other action on the same stack and for the same target sequence
is a 
negative example. The computation flow for the forward pass of
this training example is depicted in Figure~\ref{fig:tnn}.

\paragraph{Training}
The TNN then learns to predict the right action in each situation by following 
the batch gradient 
descent algorithm~\cite{DBLP:conf/kdd/LiZCS14}. Each batch consists of
all the examples derived from a particular sequence-solution pair.
To speed up the training within a batch, subprograms as well as the numbers in the 
common 
sequence are shared in the 
computation tree. 

\paragraph{Noise}
To encourage exploration, half of the searches are run with some noise added to
the (normalized) policy distribution $q$ returned by the TNN.
The noisy policy distribution $q'$ is obtained by combining a random 
distribution $r$ with $q$. 
The random distribution $r$ is created by choosing a 
real number between 0 and 1 uniformly at random for each action
and then normalizing $r$ to turn it into a probability distribution. The final 
step is to make a 
weighted average of the 
two probability distributions $q$ and $r$ to create a noisy policy $q'$ with 10 
percent noise. In the end, the noisy policy value for an action 
$a$ is $q'_a = 0.9 \times q_a + 0.1 \times r_a$.

\paragraph{Embedding integer sequences}
Integers are embedded in base 10 with the help of a digit 
concatenation operator ($::_d$) operator and a unary minus ($-_u$) operator. 
For example, the number -159 is 
represented by the tree $-_u (9 ::_d (5 ::_d 1))$.

 In order to speed up training, 
we only embed the first 16 elements of the targeted sequence.
Moreover, integers larger than $10^6$ are replaced by a 
single embedding given by a new nullary operator $\mathit{big}$. For integers 
smaller than $-10^6$ the embedding of the tree $-_u(\mathit{big})$ is computed.
With these restrictions, 91.6\% of OEIS sequences can in theory be
uniquely identified by their embeddings.

\section{Experiments}\label{sec:experiments}

Our experiments are performed on the full OEIS. The target sequence for each 
search is chosen uniformly at random from this dataset.

Each of these experiments is run on a server with 32 hyperthreading Intel(R) 
Xeon(R) CPU E5-2698 v3 @ 2.30GHz, 256 GB of memory, and no GPU cards. The 
operating system of the server is Ubuntu 20.4, GNU/Linux 5.4.0-40-generic 
x86\_64.

The code for our project is publicly available in our
repository~\cite{oeis-synthesis}. The repository contains a full list of all 
the solutions found by our algorithm during the main self-learning run.
A web interface is provided for demonstration 
purposes~\cite{oeis-synthesis-web}.
In the following, we present the chosen hyperparameters for each phase.

\subsection{Hyperparameters}
\paragraph{Search parameters}
Each search phase is run in parallel on 16 cores targeting a total of
160 targets.
On each target, a search is run for 10 minutes (2 minutes for side experiments -- cf. Section~\ref{sec:side}).
\paragraph{Test parameters}
A program $p$ is tested (evaluated) with an initial time limit of 50 microseconds.
Additional 50 microseconds are added to the timeout each time $p$
generates a new term. This means that a program $p$ times out if it takes 
more than a millisecond (= $20 \times 50$ microseconds) to generate the 
first 20 terms. The execution also stops when an integer with an absolute value 
larger than $10^{285}$ is produced.  This bound was chosen to be larger than the
largest number (in absolute value) in the OEIS, which is approximately equal to 
$4.685 \times 10^{284}$.
\paragraph{Training parameters}
The TNN embedding dimension $d$ is chosen to be 64.
Each neural network block consists of two fully-connected layers 
with $\mathit{tanh}$ activation functions.
To get the best performance, our TNNs are implemented in \xspace{C} with the 
help of the Intel MKL library~\cite{mkl}. 

The values for these parameters were optimized over multiple self-learning runs.
The variation in the number of solutions with respect to multiple parameter 
values was taken into account to select the best parameters.
These experiments were typically stopped before the 5th generation and ran
with a timeout of 2 minutes per search.

\subsection{Side experiments}
\label{sec:side}

We have identified four sources of randomness that can affect the results of 
our 
experiments:
\begin{itemize}
\item the random selection of the target sequences,
\item the initialization of the TNN with the random weights,
\item the random noise added on top of the policy in half of the searches,
\item and the fluctuations in the efficiency of the server (interacting with 
the time limits).
\end{itemize}

In our side experiment (E1), we 
measure the degree to which they influence our 
results. In particular, we run a self-learning run for 5 generations three times with
the same parameters. The differences in the numbers of solutions are
reported in Table~\ref{tab:noise}. 
The worst performing run gives $4\%$ fewer solutions than the best 
performing
run, demonstrating the robustness of the system.
\begin{table}[]
  \begin{small}
	
	\caption{Side experiments: solutions at each 
	generation.}\label{tab:noise}
	\begin{center}
	\begin{tabular}{rrrrrr}
		\toprule Generation & 0 & 1 & 2 & 3 & 4\\
		\midrule
		Experiment (E1) & 280 & 4317 & 6970 & 8724 & 
		10139\\  
		          & 259 & 4624 & 7244 & 8771 & 9767\\
		          & 310 & 4905 & 7275 & 8570 & 9747\\
		\midrule
		Experiment (E2) & 261 & 4334 & 6410 & 7614 & 8796\\
		\bottomrule
	\end{tabular}
    \end{center}
     \end{small}
    \end{table}
In our side experiment (E2), we evaluate the effect of selecting random 
solutions instead of the smallest ones during the test phase. This experiment 
is also run for 5 generations and its results are included in 
Table~\ref{tab:noise} for a comparison. 
We observe a decrease in the number of solutions by about 10 percent when 
compared with the worst performance of the default selection strategy.

Thus, we can say that selecting the smallest solutions instead of the random ones 
for training 
helps finding new solutions in later generations.
This experiment also shows how efficient this additional application of
Occam's razor is in our setting.
Note that due to the bottom-up nature of 
our tree search, smaller solutions are already more likely to be constructed 
first. Therefore the randomly selected solutions are already relatively 
small. In other words, our explicit application of  Occam's razor
(selecting the smallest solutions) is combined with an implicit bias towards Occam's razor
given by the nature of our search.


\subsection{Full-scale self-learning run}\label{sec:result}
Finally, we perform a full-scale self-learning run with a larger timeout of 10 minutes per search 
and for a larger number of generations.
The cumulative numbers of solutions found after each generation are shown in Table~\ref{ResultsTableFull}
and plotted in Figure~\ref{fig:rl}.

\begin{table*}[t]
  \caption{\label{ResultsTableFull}Solutions (second row) found after $x$-th 
  generation (first row).}
\centering
\begin{tiny}
 \begin{tabular}{llllllllllllll}
\toprule
0 &    1 &     2 &     3 &     4 &     5  &  6 &  7 &  8 & 9 & 10 & 11 & 12 \\
993 & 9431 & 13419 & 15934 & 18039 & 19412  & 20607 & 21679 & 22572 & 23284 & 23968 & 24540 & 24552 \\
\midrule 
13 &   14 & 15 & 16 & 17 & 18 &  19 &  20 & 21 &  22 &   23 & 24 &    25 \\
25006 & 25425 & 25856 & 26196 & 26509 & 26644 & 26938 & 27038 & 27041 & 27292 & 27606 & 27800 & 27987 \\
\bottomrule
  \end{tabular}
\end{tiny}
\end{table*}

\pgfplotscreateplotcyclelist{rw}
{solid, mark repeat = 10, mark phase = 200, mark = *, black\\
	solid, mark repeat = 1, mark phase = 1, mark = square*, black\\}

\begin{figure}[t]
  \centering
  \begin{minipage}[b]{\linewidth}
	\begin{tikzpicture}[scale=1.0]
	\begin{axis}[
	legend style={anchor=north east, at={(0.95,0.4)}},
	width=0.9\textwidth,
	height=0.5*\textwidth,xmin=-0.2, xmax=25.2,
	ymin=0, ymax=29000,
	cycle list name=rw,
	ytick = {0,5000,10000,15000,20000,25000,30000},
	xtick = {0,5,10,15,20,25},
	scaled y ticks = false,
	ylabel near ticks,
	xlabel near ticks,
	xlabel = {Generation},
	ylabel = {Solutions} 
	]
	\addplot table[x=gen, y=sol] {rldata};
	\end{axis}
	\end{tikzpicture}
	\caption{Solutions found after $x$-th generation\label{fig:rl}}
      \end{minipage}
      \quad
      \begin{minipage}[b]{\linewidth}
	\centering
	\begin{tikzpicture}[scale=1.0]
	\begin{axis}[
	ytick={0,1000,2000,3000},
	width=0.9*\textwidth,
	height=0.5*\textwidth,
	xmin=0, xmax=50,
	ymin=0, ymax=3200,
	cycle list name=rw,
	ylabel near ticks,
	xlabel near ticks,
	xlabel = {Size},
	ylabel = {Solutions} 
	]
	\addplot table[x=prog, y=num] {dissize};
	\end{axis}
	\end{tikzpicture}
	\caption{\label{fig:dis_size}Number of solutions with size x}
      \end{minipage}
    \end{figure}

The unguided generation 0 discovers only 993 solutions.
    After learning from them, the system discovers additional 8438 solutions in generation 1. 
There are two reasons for this effect. First, the learning 
globally biases the search towards solutions of the sequences that are in the OEIS. Second,
after the first learning, the TNN takes into account
which sequence is being targeted. Therefore each TNN-guided search explores
a different part of the program space.

At generation 25, 187 new solutions are constructed which is still more
than the 160 targeted sequences.
Thus, it is clear that hindsight experience replay is crucial in our setting.

\paragraph{General evaluation metrics}
To evaluate progress relative to other future competing 
works, we propose to judge the quality of a synthesis system based on four
criteria:
\begin{itemize}
\item The number of sequences fully generated: 27987 .
\item The average size of the solutions (smaller is better): 17.4 .
\item The simplicity of the language: 13 basic operators .
\item The computation power required: less than 100 CPU days (no GPUs were used).
\end{itemize}

We are not aware of any other self-learning 
experiment on the OEIS that provides all this information for comparison.
Further discussion of related works is provided in 
Section~\ref{sec:related}.
The distribution of programs according to their sizes is 
presented in Figure~\ref{fig:dis_size}. The long tail of this distribution is not shown in this 
figure. The largest solution has size 455.

\paragraph{Generalization to larger inputs}

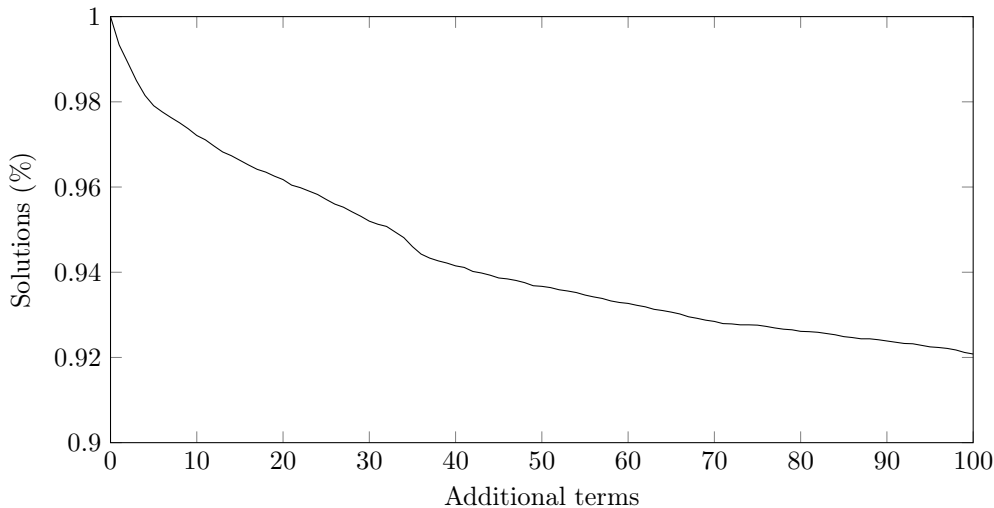
\begin{figure}[]
	\begin{tikzpicture}[scale=1.0]
	\begin{axis}[
	width=0.9*\linewidth,
	height=0.5*\linewidth,xmin=0, xmax=100,
	ymin=0.9, ymax=1,
	cycle list name=rw,
	ylabel near ticks,
	xlabel near ticks,
	xlabel = {Additional terms},
	ylabel = {Solutions (\%)} 
	]
	\addplot table[x=x, y=proba] {largeinput};
	\end{axis}
	\end{tikzpicture}
	\caption{\label{fig:largeinput}Percentage of solutions 
    that correctly generate the next $x$ additional terms.}
\end{figure}

For each sequence $\mathit{seq}$ in the OEIS, we denote by $n_{\mathit{seq}}$ the 
number of terms in the standard OEIS dataset used in our experiments.
The frequency of each sequence length in this dataset is given in 
Figure~\ref{fig:len}.
To test the generalization of our solutions to larger inputs, we look for 
additional terms for each sequence. For many OEIS 
sequences, such information can be found in b-files provided by the OEIS 
website~\cite{oeis}.
We verify further 12639 solutions that have an extra 100 
terms in their corresponding b-file with a maximum absolute value of $10^{285}$
for each term.
The proportion of sequences generated by those solutions, that covers
$x$ additional terms, is plotted in 
Figure~\ref{fig:largeinput}.  More than 92\% of the generated sequences match 
their intended counterpart on the first $n_{\mathit{seq}} + 100$ terms. This 
result confirms the 
premise of our introduction that small programs provide good explanations for 
the particular case of mathematically interesting integer sequences.

Testing alone however can not 
guarantee that our synthesized programs produce their corresponding OEIS 
sequences, as defined by their English descriptions, for all inputs $x \in 
\mathbb{N}$.
In the following analysis (Section~\ref{sec:analysis}), we will show that this is the case for a few 
selected sequences.

\arrangementX[A]{twocol}
\section{Analysis of the solutions}\label{sec:analysis}

To understand the abilities of our system, we  
analyze here some of the programs discovered by the full-scale self-learning run (Section~\ref{sec:result}).

As a measure of how important a program $p$ is for the construction of
other programs, 
we also compute the number of occurrences of $p$ in other programs 
(with 
multiple occurrences per program counted only once). This number is below indicated 
in brackets between the OEIS number (A-number) and its OEIS description.

The programs can also be searched for and analyzed using our web
interface~\cite{oeis-synthesis-web}.  Its typical use is to enter the
initial terms of a sequence, adjust the search parameters, explore the best matching OEIS sequences, and present
the programs found both in our language and after a translation to
Python. The Brython\footnoteA{\url{https://brython.info/}} interactive
editor can then be used to explore, modify and immediately run the discovered 
programs. 
For some of the solutions, we additionally write 
their expressions using recurrence relations (in the next bullet after their 
native formulations) in order to facilitate their analysis. 

\begin{enumerate}
\item \emph{Solution for A1146}\footnoteA{\url{https://oeis.org/A001146}}\ [60 
occurrences], $s_x = 2^{2^x}$.
\begin{small}
\begin{align*}
&\bullet \mloop(\lambda(x,y).x \times x,\ x,\ 2)\\
&\bullet u_x\ \mbox{where}\ u_0 = 2, u_n = u_{n-1} 
\times u_{n-1}
\end{align*}
\end{small}
Unexpectedly, only a single loop was needed to create 
this fast increasing function. 

\item \emph{Solution for A45}\footnoteA{\url{https://oeis.org/A000045}}\ [134 
occurrences], the Fibonacci sequence:
\begin{small}
\begin{align*}
&\bullet \mloopt(\lambda(x,y).x + y, \lambda(x,y).x, x, 0, 1)\\
&\bullet u_x \ \mbox{where}\ u_0 = 0, v_0 = 1, u_n = u_{n-1} + 
v_{n-1}, v_n = u_n
\end{align*}
\end{small}
This implements an efficient linear algorithm, whereas the natural 
implementation of the recurrence relation $u_n = u_{n-1} + u_{n-2}$ is 
exponential.

\item \emph{Solution for A312}\footnoteA{\url{https://oeis.org/A000312}}\ [74 
occurrences], $s_x = x^x$ :
\begin{small}
\begin{align*}
&\bullet \mloopt(\lambda(x',y').x' \times y', \lambda(x',y').y', x, 
1, x)\\
&\bullet u_x \ \mbox{where}\ u_0 = 1,\ v_0 = x,\ u_n = 
u_{n-1} \times v_{n-1},\ v_n = v_{n-1}
\end{align*}
\end{small}
This example illustrates the use of the local variable $y'$ as 
a storage for the top level variable $x$. 
The value of $y'$ (or $v_n$) is constant 
throughout the loop and equal to $x$.

\item \emph{Solution for A108}\footnoteA{\url{https://oeis.org/A108}} 
, the Catalan numbers $\binom{2x}{x}\frac{1}{x+1}$:
\begin{small}
\begin{align*}
&\bullet \mloop(\lambda(x,y).2 \times (x - x\ \mdiv\ y + x), x, 1)\ \mdiv\ (1 + 
x)\\
&\bullet u_x\  \mdiv\ (1 + x) \ \mbox{where}\  u_0 = 1 \mbox{ and}\\ 
&\hspace{23mm} u_n = 2 \times (u_{n-1} - u_{n-1}\  \mdiv\ n + u_{n-1})
\end{align*}
\end{small}
We prove by recurrence on $x$ that this solution generates the Catalan numbers.
Since the denominators are equal, we only need 
to prove 
that $u_x = \binom{2x}{x}$.
The base case is true since $1 = \binom{2 \times 0}{0}$. 
For the inductive case, we assume that $u_x = \binom{2x}{x}$ and show that:
\begin{small}
\[u_{x+1} = 2(u_{x} - \frac{u_{x}}{x+1} 
+ u_{x}) = \frac{2(2x+1)} {x + 1} u_x =\]
 \[\frac{2(2x+1)}{x+1}\frac{(2x)!}{x! ^ 2} = 2(x+1)\frac{(2x+1)!}{(x+1)!^2} =\] 
 \[\frac{2(x+1)}{2x+2} \binom{2x+2}{x+1} = \binom{2(x+1)}{x+1}\]
\end{small}
\item \emph{Solution for A10051}\footnoteA{\url{https://oeis.org/A010051}} [28 
occurences], the characteristic function of the primes: 
\begin{small}
\begin{align*}
&\bullet (\mloop(\lambda(x,y).x \times y, x, x)\ \mmod\ (1 + x))\ 
\mmod\ 2\\
&\bullet (u_x \ \mmod\ (1 + x))\ 
\mmod\ 2\ \mbox{where}\ u_0 = x, u_n = u_{n-1} \times n
\end{align*}
\end{small}
We can now prove the following conjecture relating the discovered and 
reformulated formula with the prime characteristic function $1_P$: 
\begin{small}
\[\forall x \in \mathbb{N}.\ ((x \times x!)  \ \mmod\ (1+x)) \ \mmod\ 2 = 
1_P(x) \]
\end{small}
This conjecture is a variation of Wilson's theorem and
its proof reveals how the generated formula is structured. 
We reason modulo $1+x$, saying that $a \equiv b$ if $a$ and $b$ are equal modulo 
$1+x$. We prove the conjecture by considering four cases:
\begin{itemize}
\item If $1+x$ is prime, then every non-zero element has an inverse in the 
$Z_{1+x}$ 
and the only elements that are their own inverses are $1$ and $-1 \equiv x$. 
Indeed, a 
field has a maximum of two solutions for the equation $X^2 \equiv 1$ according 
to the 
fundamental theorem of algebra.
Thus, the elements of the product $x!$ can be regrouped into pairs of inverses 
except for $x$, hence $x \times x! \equiv x \times x \equiv 1$.
\item If $1+x$ is not prime and is not the square of a prime, then $x!$ divides 
$1+x$ 
since $1+x$ has two distinct proper divisors. 
\item If $1+x=p^2$ where $p$ is an odd prime, then $(p^2-1)!$ divides 
$p^2$ since $p$ and $2p$ appear in $(p^2-1)!$.
\item If $1+x=4$, then $3 \times 6 \equiv 2$. Only in this case, the final 
modulo 2 operation is necessary.
\end{itemize}
	
From this primality test, the operator $\mcompr$ can construct 
the set of prime numbers. However, this construction fails to pass the testing 
phase because it is too inefficient.
The creation of an efficient prime number generator is an open problem at the 
intersection of mathematics and computer science that we were not expecting to 
solve.
In a previous run, a slightly more efficient but inaccurate way of generating 
prime numbers was re-discovered by synthesizing the Fermat primality test in 
base 2. The generated sequence of pseudoprimes\footnoteA{\url{https://oeis.org/A015919}}
deviates from the prime number sequence. Indeed,
the number 341 passes Fermat's primality test but is not prime.

\item \emph{Solution for A46859},\footnoteA{\url{https://oeis.org/A046859}} 1 3 
7 61, ``simplified'' Ackermann:
\begin{small}
\[\mloop(\lambda(x,y).\mloopt(\lambda(x,y).x + y, \lambda(x,y).x, x, 2, 2) - x, 
x, 1)\]
\end{small}
Our pseudo-solution is primitive recursive while Ackermann is not. 
The next number this solution generates
is 13114940639623 shy of the expected $2^{2^{2^{2^{16}}}} - 3$ 
too large to be included in the OEIS. Generally, our system 
is likely to give the ``wrong'' solution when sequences have very few terms.

\item \emph{Solution for A272298},\footnoteA{\url{https://oeis.org/A272298}} 
$s_x = x^4 + 324$:
\begin{small}
\[\mloop(\lambda(x,y).x \times x, 2, x) + 1 + 1 + 1 + \ldots + 1\]
\end{small}
This is the longest solution generated by our system. 
It has size 654 as
the number 324 is constructed by repeatedly adding $1$.
A shorter way of expressing the constant $324$ is found before the 
end of our self-learning run:
\begin{small}
  \[\mloop(\lambda(x,y).x \times x, 2, 1 + 2) \times (2 + 2) = 3^4 \times 4\]
\end{small}

\item \emph{Solution for A66298},\footnoteA{\url{https://oeis.org/A066298}} 
$s_x = \mathit{googol}\ (\mathit{mod}\ x)$;
\begin{small}
\[\mloop(\lambda(x,y).\mloop(\lambda(x,y).(2 + (2 \times (2 + 2))) \times x, x, 
1), 2, 2)\] \[ \mmod\ (1 + x)\]
\end{small}
It contains the largest constant $10^{100}$ used in our solutions. After 
recognizing the sub-expression for $10^x$, this program can be rewritten as 
$10^{10^2}\ \mmod\ (1 + x)$. This program uses $1+x$ instead of $x$ since the 
sequence starts at $s_1$ in the OEIS.

\item Solution for A195,\footnoteA{\url{https://oeis.org/A000195}} $s_x = \mathit{floor}(\mathit{log}(x))$:
\begin{small}
\[((\mloop(\lambda(x,y).(y\ \mdiv\ \mloopt(\lambda(x,y).x + y, \lambda(x,y).x, 
x, 1, 1))\]
\[ + x, 1 + x, 1) - 1)\ \mmod\ (1 + x))\ \mdiv\ 2\]
\end{small}
Among the 133 solutions found at generation 18, this is the 
smallest. It has size 25.

Further work is required to check if the last two presented solutions are 
correct. In the future, we will aim to prove that the 
intended solution and the synthesized one are equal with the help of 
interactive and automatic theorem provers. 
Solutions discovered during the full self-learning run are available for 
further analysis in our repository~\cite{oeis-synthesis}.
\end{enumerate}

\section{Related work}\label{sec:related}
The closest related recent work is~\cite{DBLP:journals/corr/abs-2201-04600},
done in parallel to our project.
The goal there is similar to ours but their 
general approach is focused on training a single 
model
using supervised learning techniques on synthetic data.
In contrast, 
our approach is based on reinforcement learning: we start from scratch and keep training new models 
as we discover new OEIS solutions.

The programs generated in~\cite{DBLP:journals/corr/abs-2201-04600} are 
recurrence relations defined by analytical formulas.
Our language seems to be more expressive. For example, our language can use the 
functions it defines by recurrence using $\mloop$ and $\mloopt$ as subprograms 
and construct nested loops.
To fit the Transformer architecture, \cite{DBLP:journals/corr/abs-2201-04600} represent programs as a sequence of 
tokens, whereas our tree neural network does not need such transformations.
The performance of the model in ~\cite{DBLP:journals/corr/abs-2201-04600} is investigated on real 
number sequences, whereas our work focuses only on integer sequences. 
Overall, it is hard to directly compare the performance of the two systems.
Our result is the number of OEIS sequences found by targeting the 
whole encyclopedia, whereas~\cite{DBLP:journals/corr/abs-2201-04600} report the test accuracy only on 10,000 easy OEIS 
sequences.

The inspiration for many parameters of our self-learning loop comes principally 
from AlphaGoZero~\cite{silver2017mastering} which uses 
self-play
to learn  Go at a professional level and above. The main difference in our 
setting is that synthesis is essentially a one-player game.
Therefore, rewards are sparse as the difficulty of the 
problems does not match the level of understanding of our system.
That is why we use a form of \textit{hindsight experience replay} to increase 
the number 
of training examples.

Finally,
the deep reinforcement learning system 
\textsf{DreamCoder\xspace}~\cite{DBLP:conf/pldi/EllisWNSMHCST21} has 
demonstrated 
self-improvement from scratch in
various programming tasks. Its main contribution
is the use of 
definitions
that compress existing solutions and facilitate building new
solutions on top of the existing ones. As seen from the experiments 
in~\cite{DBLP:conf/pldi/EllisWNSMHCST21}, adding definitions
typically works up to a point. After that point, the extra actions are 
hurting significantly the chance of constructing a program that does not need 
them.

\section{Conclusion}
Our system has created from scratch programs that generate
the full list of available terms in the OEIS for 27987 sequences.
Based on \emph{Occam's razor}, we have argued that producing 
shorter programs is more likely to generate better
explanations for particular sequences. We have also shown that 
the solutions discovered are correct for some famous sequences.
And we have observed that preferring shorter programs during the training 
increases the performance of the system.

In the future, we would like to create a benchmark of theorem proving problems 
from the solutions found in our experiments.
There, automatic theorem provers would be tasked to
prove that a particular intended definition for a sequence (e.g. the prime 
characteristic function $1_P(x)$) is equivalent to the 
program that we have discovered (e.g. $((x \times x!)  \ \mmod\ (1+x)) \ 
\mmod\  2$). We believe that such a 
non-synthetic benchmark would contribute greatly 
to the development of automated inductive reasoning.

\section*{Acknowledments}
This work was partially supported by the CTU Global Postdoc funding
scheme (TG), Czech Science Foundation project 20-06390Y (TG), Amazon
Research Awards (TG, JU), EU ICT-48 2020 project TAILOR no. 952215 (JU),
and the European Regional Development Fund under the Czech project
AI\&Reasoning no. CZ.02.1.01/0.0/0.0/15\_003/0000466 (JU).

\bibliographystyle{plain}
\bibliography{biblio}
\end{document}